\documentclass[11pt,a4paper]{article}
\usepackage[hyperref]{acl}
\usepackage{times}
\usepackage{latexsym}
\usepackage{amsmath}
\usepackage{amssymb}
\usepackage{booktabs}
\usepackage{graphicx}
\usepackage{multirow}
\usepackage{url}

\usepackage{amsmath}
\usepackage{tikz}
\usetikzlibrary{calc, positioning, arrows.meta, shapes.geometric, backgrounds, shadows}

\newcommand{\commentOut}[1]{}

\title{Interpretable-by-Design Transformers via \\
Architectural Stream Independence}

\newcommand{\tokpos}{{token position}}
\newcommand{\tokstream}{{token stream}}
\newcommand{\symstruct}{{symbolic structure}}
\newcommand{\symhead}{{symbolic head}}
\newcommand{\poshead}{{position head}}
\newcommand{\rechead}{{recency head}}
\newcommand{\PDS}{{Token-Position Dependence Score (PDS)}}

\author{Clayton Kerce \thanks{This work was partially supported by contract HR001125C0302} \\ Georgia Tech Research Institute \\ \texttt{clayton.kerce@gtri.gatech.edu} \and \\ Alexis Fox \\ Georgia Tech Research Institute \\ \texttt{alexis.fox@duke.edu}}

\begin{document}
\maketitle

\begin{abstract}
While transformers achieve strong performance, their internal
decision-making processes remain opaque. We investigate whether
architectural constraints can enforce \textit{interpretability by design}
through \textbf{architectural stream independence}: maintaining
a \textbf{\tokstream}\ (carrying \symstruct) and \textbf{contextual semantics} in separated streams that remain independently
observable throughout processing, with integration delayed until output.

We validate this principle through the Late Fusion Architecture (LFA),
which demonstrates \textbf{interpretable \symhead s through all the final layers}, while standard transformers show dissolution by the third of six layers; we quantify this effect by introducing the \PDS, with PDS$_{max} = $ 0.276 and 0.058, respectively. Crucially, intervention experiments demonstrate functional modularity: suppressing LFA's \rechead s causes minimal semantic damage (Cohen's $d = -0.158$) versus catastrophic entanglement in baselines ($d = -0.672$).

LFA demonstrates that architectural constraints improve underlying learning 
mechanisms, averaging 42\% stability versus 19\% and 11\% for baseline 
comparisons, with extremes from 50\% on LFA's best pairs (6 of 12 heads 
position-invariant) down to 0\% complete collapse in over-constrained cases. By preventing premature entanglement, architectural independence steers models toward semantic understanding over positional heuristics, establishing interpretability as an architectural design criterion enforceable through structural constraints rather than post-hoc analysis.
\end{abstract}


\section{Introduction}

Despite the success of transformer-based language models, their internal mechanisms remain largely opaque. When models exhibit failures, e.g. recency bias \cite{guo2024serialpositioneffectslarge}, sycophancy \cite{sharma2024understandingsycophancylanguagemodels}, or spurious correlations \cite{yang2024identifyingspuriousbiasesearly}, practitioners lack tools for understanding and addressing root causes. Post-hoc interpretability methods \cite{belinkov2017neural,clark2019does} reveal what models learn or where they attend, but do not provide paths toward models that are interpretable \textit{by design}.

It is natural to ask: \textit{Can we identify specific mechanisms within a model's internal activations that control high-level behaviors?} This work directly addresses a strong form of explainability: rather than analyzing what emerged during training, can we design architectures that are more interpretable by construction? We answer affirmatively by demonstrating that \textbf{architectural constraints can promote transparent reasoning pathways} whose internal mechanisms are modular and independently observable.

To this end, we test a specific architectural hypothesis: \textbf{architectural stream
independence}---maintaining a \textit{\tokstream}\ and \textit{contextual semantics} in separated streams that
remain independently observable throughout processing---preserves functional
modularity, creating models whose mechanisms can be directly observed and
independently intervened upon. This contrasts with immediate integration
architectures that immediately mix \symstruct\ and contextual information, causing \symstruct\
signals to dissolve into entangled representations.

This design is motivated by intuitions about distinct computational roles---symbolic routing via attention, pattern discovery through circuit formation,
and contextual elaboration via the feed-forward network (Appendix~\ref{app:reasoning_types})---
that benefit from architectural separation. We implement architectural stream
independence through \textbf{asymmetric information flow}: a frozen position
stream ($X_T$, carrying \symstruct) influences {contextual} updates ($X_E$) without being corrupted
by gradient flow, with symmetric combination only at the output layer (lm\_head).
Unlike DeBERTa's additive decomposition within attention \cite{he2021deberta},
our approach maintains independent observability: \symstruct\ remains a clean,
surgically analyzable signal throughout all transformer layers.

\paragraph{Notation.}
We distinguish between the \textbf{absolute token index} $i \in \{0, \dots, T-1\}$ and the \textbf{relative token position} $p = i_{\text{target}} - i$ (used in intervention analysis). We denote \textit{layer index} by $\ell$ and \textit{head index} by $h$. The model maintains two parallel streams: a frozen \tokstream\ $X_T$ encoding \symstruct, and a mutable {contextual stream} $X_E$.

We validate this through the Late Fusion Architecture (LFA), demonstrating: \textbf{(1)} functional transparency as a design criterion with quantitative metrics; \textbf{(2)} systematic comparison of four architectural variants isolating constraint effects; and \textbf{(3)} empirical validation showing LFA maintains interpretable position channels in layers 4-5 (PDS$_{max}$ = 0.276) where immediate integration architectures show premature dissolution (PDS$_{max}$ = 0.058), with measurably lower intervention collateral damage (Cohen's $d = -0.158$ vs. $-0.672$ for entangled baselines).

Our coreference experiments show that LFA's specialists concentrate in
mid-to-late layers (top head L4.H3: 48.3\%), while standard transformers'
best heads are distributed across layers (best head L1.H5: 46.9\%;
Table~\ref{tab:coreference}).
We validate using small models (13M-22M parameters) on TinyStories,
providing clean evidence that \textbf{interpretability can be designed into architectures through principled constraints},
offering a path toward transparent, explainable language models where
internal reasoning processes can be directly observed and understood.

\section{Architecture and Design Principles}

\subsection{Architectural Stream Independence: The Design Principle}

We define \textbf{architectural stream independence} as maintaining
a \tokstream\ and \textbf{contextual semantics} in separated streams that remain independently
observable throughout transformer processing, with integration occurring
only at the output layer. This design principle creates transparent
reasoning pathways where distinct mechanisms can be directly observed
and independently intervened upon.

Standard transformers violate this principle through \textbf{immediate
integration}: \tokpos\ encodings and token identities are added at layer 0 and immediately
mixed with semantic features via dense attention.
By layer 2, \symstruct\ dissolves into distributed semantic representations, making it impossible to isolate which heads track \textbf{symbolic identities} versus meaning
(Figure~\ref{fig:pds_layer}, Std-T panel).

\textbf{The Late Fusion Architecture (LFA)} implements architectural
stream independence through delayed integration: \textit{\tokstream}\ and \textit{contextual semantics}
remain separated until output-layer combination. This ``late fusion''
timing preserves functional modularity---by keeping streams independent
until the model has developed rich semantic representations, we create
mechanisms that remain independently observable and interventionable.

This design generates four testable predictions:
\begin{enumerate}
\item \textbf{Preserved observability:} LFA maintains independently observable \textbf{\symhead s} in deep layers (L4-L5), validated by high PDS
\item \textbf{Early dissolution:} Immediate integration architectures show
\symstruct\ dissolving by layer 2, validated by low PDS
\item \textbf{Functional independence:} Separated streams enable surgical
interventions on \textbf{\rechead s} without semantic damage
\item \textbf{Entangled opacity:} Immediate integration creates inseparable
mechanisms where interventions cause catastrophic collateral damage
\end{enumerate}

\subsection{Stream Separation Mechanism}

We implement architectural stream independence through \textbf{separation of streams}.
The model maintains:
\begin{itemize}
\item \textbf{Frozen \tokstream\ ($X_T$):} preserves \symstruct\ and \tokpos\ unchanged
\item \textbf{Embedding stream ($X_E$):} accumulates semantic updates
\end{itemize}

Formally:
\begin{equation}
\begin{aligned}
X_T^{(0)} &= \text{TokenEmbedding}(\text{input}), \quad X_E^{(0)} = 0 \\
X_T^{(\ell)} &= X_T^{(0)} \quad \text{(frozen across all layers)}
\end{aligned}
\end{equation}
The embedding stream updates via:
\begin{equation}
\begin{aligned}
X_E^{(\ell)} &= X_E^{(\ell-1)} + \text{Attn}(Q, K, X_T^{(0)}) \\
&\quad + \text{FFN}(X_T + X_E)
\end{aligned}
\end{equation}

Critically, while the FFN incorporates positional information into semantic
processing (observing $X_T + X_E$), it maintains architectural separation by
writing only to $X_E$. This \textbf{asymmetric information flow} allows
\tokpos\ and \symstruct\ to inform semantic learning while preserving the \tokstream\ as a clean,
independently observable signal. Attention similarly reads from both streams
but writes only to $X_E$.

The streams remain separated until lm\_head combines them for final prediction:
\begin{equation}
\text{logits} = \text{LM\_head}(\text{LayerNorm}(X_T + X_E))
\end{equation}

This delayed \textbf{symmetric combination} is the ``late'' in late fusion---
\symstruct\ remains independently observable throughout all internal processing,
with no gradient flow corrupting $X_T$.

\subsection{Distinction from Prior Disentanglement}

DeBERTa computes separate position-content attention terms but allows
immediate mixing \cite{he2021deberta}. Our frozen stream enforces
\textbf{architectural separation} with no gradient flow between
streams. DeBERTa disentangles \textit{how} attention is computed;
we control \textit{where} updates are written. This architectural
constraint, combined with independent channels (ChannelLayerNorm +
Kronecker mixing), maintains complete separation throughout all
transformer layers until the final output.

\subsection{Why Asymmetric Flow Maintains Architectural Independence}

The asymmetric information flow ($X_T \rightarrow X_E$, but $X_T$ frozen)
is critical for maintaining architectural stream independence:

\begin{enumerate}
\item \textbf{Position never corrupted:} $X_T$ receives no gradient updates,
remaining a clean position signal
\item \textbf{Semantics benefit from position:} FFN observes $X_T + X_E$,
enabling position-aware semantic processing
\item \textbf{Independent observability preserved:} Both streams remain
surgically analyzable and interventionable
\item \textbf{Delayed integration:} Symmetric combination (lm\_head) occurs
between independently-evolved, mature representations
\end{enumerate}

High PDS in deep layers validates that architectural independence is maintained: position signals remain distinct and independently observable. If premature integration occurred (as in standard transformers), position would dissolve into distributed semantic representations, destroying independent observability.

Table~\ref{tab:early_late_fusion} summarizes the architectural contrast between early and late fusion.

\begin{table}[t]
\centering
\small
\begin{tabular}{lcc}
\toprule
Property & Immediate & Stream \\
 & Integration & Independence \\
 & (Std-T) & (LFA) \\
\midrule
Position encoding & Added at L0 & Frozen in $X_T$ \\
Stream separation & None & $X_T$ / $X_E$ distinct \\
Information flow & Bidirectional & $X_T \rightarrow X_E$ only \\
Position observable L5 & No (PDS=0.058) & Yes (PDS=0.276) \\
Integration timing & Every layer & lm\_head only \\
Gradient flow to $X_T$ & Yes & No \\
\bottomrule
\end{tabular}
\caption{\textbf{Immediate integration vs. architectural stream independence.}
LFA maintains independent observability through asymmetric information flow
and delayed symmetric combination.}
\label{tab:early_late_fusion}
\end{table}

\subsection{Architectural Constraints}

\textbf{Frozen Token Stream (FTS):} Token embeddings initialize $X_T^{(0)}$, while $X_E^{(0)}=0$. In FTS, attention reads from the frozen token stream and writes to the embedding stream; $X_T$ is held fixed across layers while all parameters train end-to-end. Channelization is preserved by construction (identity $W_V$ and $W_O$ in FTS). Complete stream separation is maintained throughout all transformer layers; fusion occurs only at the output layer (lm\_head). See Appendix~\ref{app:architecture} for the formal update rules.

\subsection{Model Configurations}

We train four models to test our hypothesis and ablate architectural components (Table~\ref{tab:training}):

\textbf{LFA (Late Fusion Architecture):} Implements architectural stream
independence via FTS + independent attention + dense FFN. Maintains separated
streams with integration delayed until output.

\textbf{Std-T (Standard Transformer):} Single-stream with dense attention and
FFN (GPT-2 style). Immediate integration baseline.

\textbf{D-Cas (Dense Attention Baseline):} FTS + dense attention + dense FFN.
Tests whether frozen stream alone improves interpretability.

\textbf{CFM (Channel-Factored Model):} FTS + independent attention +
independent FFN. Tests whether excessive constraint degrades learning.

All models use 6 layers, 6 heads, trained on TinyStories for 2 epochs. Model sizes vary by constraint (13.4M-22.2M parameters). The dense FFN in LFA serves a dual computational role: (1) it observes both streams to inform semantic updates, enabling rich contextual processing while (2) writing only to $X_E$, maintaining architectural separation. Unlike CFM's independent FFN which cannot coordinate across heads, LFA's dense FFN enables semantic coordination while preserving channel independence in attention. Critically, fusion occurs only at lm\_head, combining the frozen \tokstream\ ($X_T$) with matured semantic representations ($X_E$). While our experiments focus on validating the interpretability benefits of architectural separation, the computational motivations behind these design choices are discussed in Appendix~\ref{app:reasoning_types}.

\begin{table}[t]
\centering
\small
\begin{tabular}{llrrr}
\toprule
Model & Attn & FFN & Params & Val Loss \\
\midrule
Std-T & dns & dns & 22.2M & 1.8114 \\
D-Cas & dns & dns & 21.2M & 1.8399 \\
LFA & ind & dns & 18.5M & 1.9063 \\
CFM & ind & ind & 13.4M & 2.1019 \\
\bottomrule
\end{tabular}
\caption{Model configurations and training performance. All models have 6 layers, 6 heads, 384d embeddings, trained on 2M samples from TinyStories for 2 epochs.}
\label{tab:training}
\end{table}

\section{Explainability Methodology: Isolating Positional from Semantic Reasoning}

\subsection{Coreference Resolution Analysis}

We construct a ground truth dataset of 29 coreference instances across 13 diagnostic prompts, each specifying a pronoun (query token) and correct antecedent. The dataset includes minimal pairs testing specific phenomena: competing nouns (semantic appropriateness vs. positional recency), gender resolution, and plurality agreement. For example, competing noun pairs test whether models attend to semantically appropriate targets regardless of position: ``Tim saw a \textit{key} and a box. He used it'' vs. ``Tim saw a box and a \textit{key}. He used it.''

\textbf{Example behavior:} Consider the minimal pair testing semantic appropriateness
versus positional recency. For ``Tim saw a box and a key. He used it'' (key first),
LFA's top coreference head (L4.H3) attends to ``key'' with weight 0.303. When positions
reverse (``Tim saw a key and a box. He used it'', key last), the same head increases
attention to ``key'' (0.409), demonstrating \textbf{\tokpos-invariant} semantic tracking---it
identifies the semantically appropriate tool regardless of position.

Std-T lacks such specialization. Its best performing head (L3.H2) shows weaker
attention to the correct antecedent in both positions (0.226 and 0.324), and the
stability metric for this minimal pair shows the architectural difference: LFA
has 0.50 stability (6 stable heads across the pair), while Std-T has 0.25
stability (1 stable head). CFM's 0\% stability (no consistent heads) reveals
complete failure to develop semantic specialists, demonstrating pure recency bias.

We define three metrics quantifying attention patterns. \textbf{Mean Attention Weight} measures average attention mass from query to correct antecedent:
\begin{equation}
\text{Attn}_{\text{mean}} = \frac{1}{N} \sum_{i=1}^{N} \alpha_{q \rightarrow t}^{(i)}
\end{equation}

\textbf{Top1 Accuracy} is the percentage where the correct antecedent receives highest attention:
\begin{equation}
\text{Top1} = \frac{|\{i : \arg\max_j \alpha_{q \rightarrow j}^{(i)} = t\}|}{N} \times 100\%
\end{equation}

\textbf{Stability} measures consistent target preference across position variations:
\begin{equation}
\text{Stability} = \frac{|\{\text{heads}: \text{target}_{\text{first}} = \text{target}_{\text{last}}\}|}{|\text{total heads}|} \times 100\%
\end{equation}

See Appendix~\ref{app:dataset} for complete dataset specification and examples.

\subsection{Token-Position Dependence Score}

To measure whether architectural separation successfully preserves \symstruct\ and \tokpos\
as an independently observable signal, we introduce the \PDS. High PDS indicates position remains distinct (maintained stream
independence); low PDS indicates position has dissolved into semantic
representations (premature integration). The delayed integration timing in
LFA refers to \textit{when \symstruct\ ceases to be independently observable}:
LFA maintains observability throughout all layers, integrating only at output.

For each head and minimal pair (target first vs. last):
\begin{equation}
\text{PDS}_{(\ell, h)} = \left| \frac{1}{K} \sum_{k=1}^{K} \alpha_{q \rightarrow t_{\text{last}}}^{(k)} - \frac{1}{K} \sum_{k=1}^{K} \alpha_{q \rightarrow t_{\text{first}}}^{(k)} \right|
\end{equation}
where $K = 29$ minimal pairs. We use threshold PDS $> 0.075$ ($\mu$ + $\sigma$
for Std-T) for descriptive counts, but interventions use ranked top-$k$
selection. See Appendix~\ref{app:pds} for threshold justification and full
distributions.

\subsection{Measuring Functional Transparency via Intervention}

To validate that \textbf{positional} mechanisms are functionally independent from semantic processing, we perform targeted lesion studies using \textit{relative token position} ($p$). For each model, we rank heads by PDS, select top-$k$ \textit{\rechead s} ($k \in \{1,2,3,5\}$), and apply soft gating to outputs ($g \in \{1.0, 0.75, 0.5, 0.25, 0.0\}$). We measure impact using Semantic Preference Score (SPS) on 60 competing-noun prompts:
\begin{equation}
\text{SPS} = \frac{1}{M} \sum_{m=1}^{M} \left( \alpha_{q \rightarrow \text{semantic}}^{(m)} - \alpha_{q \rightarrow \text{distractor}}^{(m)} \right)
\label{eq:sps}
\end{equation}
where $M = 58$ (after filtering), semantic = tool, distractor = container.

We quantify functional independence using Cohen's $d$ effect size
(Eq.~\ref{eq:cohen_d}), where small $|d|$ near zero indicates
transparent functional decomposition. We include matched-random
suppression, semantic suppression (bottom-$k$ PDS), and baseline
controls. See Appendix~\ref{app:intervention} for complete gating
protocol, soft-suppression curves, and per-category analyses.
\begin{equation}
d = \frac{\mu_{\text{intervention}} - \mu_{\text{baseline}}}{\sigma_{\text{pooled}}}
\label{eq:cohen_d}
\end{equation}


\section{Transparent Functional Decomposition}

Architectural stream independence preserves functional modularity by preventing immediate entanglement of \textit{\symstruct}\ and \textit{context}. We test this via coreference resolution, which requires entity tracking across \tokpos s without positional confounds. \textbf{Predictions:} (1) LFA shows concentrated specialization (few strong heads), (2) Std-T shows diffuse patterns (many weak heads), (3) heads are \textbf{\tokpos-invariant}. Section~\ref{sec:pds} validates prediction (1), Section~4.2 validates (3).

\subsection{Coreference Head Specialization}

LFA exhibits strong head specialization concentrated in identifiable locations. The top head (L4.H3) correctly resolves 48.3\% of all instances with mean attention 0.323; co-primary head L3.H5 achieves identical coverage (48.3\%) with mean 0.265. These heads concentrate in mid-to-late layers (L3-L4), forming a coherent, easily identifiable processing module.

Std-T develops specialists with comparable peak performance (best head L1.H5: 46.9\%), but distributes them diffusely across layers. Table~\ref{tab:coreference} compares LFA's top coreference heads to the same head positions in Std-T, revealing the architectural difference: LFA's concentration in L3-L4 versus Std-T's distribution across L1, L3, and other layers. This concentration enables direct observation---when analyzing coreference in LFA, one examines L4.H3; in Std-T, specialists must be located through search across all 36 heads. LFA's top 5 heads all exceed 37\% Top1 accuracy; those same head positions in Std-T show weak, inconsistent patterns, with most below 15\% (see Appendix~\ref{app:coreference_full} for complete results across all 36 heads).

\begin{table}[t]
\centering
\small
\begin{tabular}{lrrrr}
\toprule
\multirow{2}{*}{Head} & \multicolumn{2}{c}{Mean Attn} & \multicolumn{2}{c}{Top1 Acc} \\
\cmidrule(lr){2-3} \cmidrule(lr){4-5}
& LFA & Std-T & LFA & Std-T \\
\midrule
L4.H3 & 0.323 & 0.033 & 48.3\% & 6.9\% \\
L3.H5 & 0.265 & 0.047 & 48.3\% & 6.9\% \\
L4.H0 & 0.211 & 0.045 & 44.1\% & 13.1\% \\
L4.H1 & 0.286 & 0.060 & 37.2\% & 13.1\% \\
L3.H0 & 0.225 & 0.049 & 46.9\% & 4.8\% \\
\bottomrule
\end{tabular}
\caption{\textbf{Coreference specialization: LFA vs. Std-T.} LFA's top 5 coreference heads (ranked by mean attention) compared to the same head positions in Std-T. LFA concentrates specialists in L3-L4; these same positions in Std-T show weak, diffuse patterns. Std-T's actual best heads (L1.H5: 46.9\%, L3.H2: 44.8\%) are distributed across different layers, requiring search to locate.}
\label{tab:coreference}
\end{table}

\subsection{Token-Position-Invariant Semantic Understanding}

The competing nouns minimal pair tests whether models prefer semantically
appropriate targets over positional recency. For \textit{``Tim saw a key/box.
He used it''} vs. \textit{``Tim saw a box/key. He used it''}, robust models
should attend to \textit{key} (the usable tool) regardless of position.

Across the competing-noun minimal pairs, LFA shows higher stability: mean
stability 0.42 (range 0.27--0.50; 3--6 stable heads per pair). Std-T averages
0.19 (range 0.125--0.25; 1 stable head per pair), while CFM averages 0.11
(range 0--0.33). The range extremes reveal architectural failure modes: LFA
achieves 50\% stability on two of three pairs, with 6 of 12 heads maintaining
semantic preference regardless of \textit{relative token position} (L4.H3, L3.H5, L4.H0, L3.H0, L4.H2,
L3.H4), demonstrating that stream separation enables robust semantic learning.
In contrast, CFM exhibits complete collapse (0\% stability) on two of three
pairs, where excessive architectural constraint prevents \textit{any} head from
learning position-invariant preferences, forcing pure recency bias. This
demonstrates that architectural design determines not just typical behavior
(42\% vs 11\% mean stability) but also failure modes: stream independence
enables semantic understanding, while over-constraint can prevent it entirely.
See Appendix~\ref{app:stability} for per-head stability analysis.

\subsection{Quantitative Validation}

The architectural difference is statistically robust: LFA's top 5 specialists
co-locate in L3-L4 and all exceed 37\% Top1 accuracy with mean attention above
0.21. For those same five head positions, Std-T averages 0.05 mean attention and
9\% Top1 accuracy, demonstrating that LFA's architectural constraints create
\textit{predictable, concentrated specialization} rather than distributed patterns
requiring exhaustive search. This localization enables transparent analysis---
practitioners can directly examine L4.H3 for coreference behavior---while
maintaining comparable peak performance to Std-T's best distributed heads
(see Appendix~\ref{app:coreference_full} for complete head-wise results).


\section{Validating Explainability Through Intervention Analysis}

\subsection{Token-Position Dependence Analysis}
\label{sec:pds}

\begin{figure}[t]
\centering
\includegraphics[width=\columnwidth]{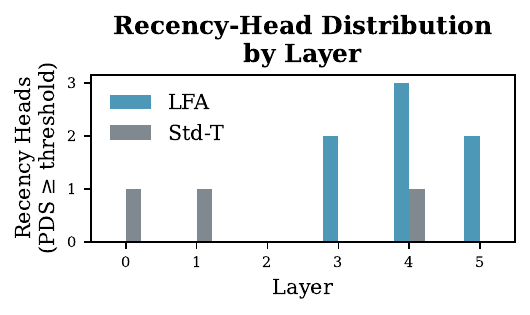}
\caption{\textbf{Temporal validation of architectural stream independence.}
LFA concentrates \textbf{positional} processing in layers 4-5, maintaining complete
stream separation throughout all transformer layers. Std-T processes position
in layers 0-1, causing immediate entanglement and dissolution by mid-layers.
CFM's mid-layer concentration reflects failed integration---excessive constraint
prevents late-layer coordination. This layer distribution directly validates
the architectural hypothesis: separated streams
preserve distinct \textbf{symbolic} channels in deep layers.}
\label{fig:pds_layer}
\end{figure}

\textbf{Temporal validation:} Figure~\ref{fig:pds_layer} validates
our design hypothesis by showing \textit{when} \textit{positional} processing
occurs. LFA maintains 5 \textit{\tokpos-dependent} heads in layers 4-5
(architectural independence preserved throughout), while Std-T
concentrates 2-3 heads in layers 0-1 with zero by layer 5 (immediate
integration followed by dissolution). This is not merely ``different
layers''---it demonstrates that LFA \textit{maintains separated streams}
throughout all transformer layers, allowing both position and semantics
to develop independently, while Std-T immediately entangles position
with under-developed features.

\textbf{Why architectural stream independence matters:} The distinction
between immediate integration (Std-T) and delayed integration (LFA) is
not merely that they use ``different layers.'' Immediate integration forces
\symstruct\ and \tokpos\ to mix with under-developed semantic representations at layer 0---when
the model has no contextual understanding yet. This immediate entanglement
means position becomes correlated noise that cannot be surgically removed
(§5.2). LFA maintains complete separation until the output layer (lm\_head),
allowing position and semantics to evolve independently through all 6
transformer layers. High PDS at layers 4-5 validates this separation is
maintained: \tokpos\ signals remain distinct and independently observable
even in deep layers, unlike Std-T where immediate mixing causes position
to dissolve by layer 2. The architecture never fuses streams internally---only
at the final output does lm\_head combine $X_T + X_E$ for prediction. This
architectural difference explains why LFA interventions cause minimal
collateral damage while Std-T interventions are catastrophic.

Our hypothesis predicts LFA maintains distinct position channels in deep
layers (L4-L5) while Std-T's immediate integration causes position signals
to dissolve. Using threshold PDS $> 0.075$ for counts, Table~\ref{tab:pds}
confirms this: LFA has {5 \rechead s in L4-L5} (max PDS = 0.276
at L5.H0), Std-T has {0 in L5} (max = 0.058), and CFM has {0}
in L4-L5 (max = 0.032).

\begin{table}[t]
\centering
\small
\begin{tabular}{lcccc}
\toprule
Model & Total & L4-L5 & Max L5 & Avg PDS \\
      & PDS$>$0.075 & Heads & PDS & \\
\midrule
LFA   & 7 & \textbf{5} & \textbf{0.276} & 0.061 \\
Std-T & 3 & 1 & 0.058 & 0.032 \\
CFM   & 5 & 0 & 0.032 & 0.042 \\
\bottomrule
\end{tabular}
\caption{\textbf{Token-Position dependence by layer.} LFA maintains distinct \textbf{symbolic} channels in deep layers (5 of 7 heads in L4-L5, max L5 PDS = 0.276), validating architectural independence is maintained. Std-T dissolves by L5 (max = 0.058), CFM fails to integrate (max = 0.032).}
\label{tab:pds}
\end{table}

Figure~\ref{fig:pds_layer} shows layer-wise distribution. LFA's deep-layer concentration (5 of 7 heads in L4-L5) validates architectural stream independence: positional information preserved in distinct channels until final layers. Std-T's early-layer concentration (2 of 3 heads in L0-L1) confirms immediate integration, with position information metabolized into semantic representations by layer 2. LFA's maximum layer-5 PDS (0.276) is \textit{5× higher} than Std-T (0.058) and \textit{9× higher} than CFM (0.032), validating architectural constraint success.

\subsection{Functional Transparency via Intervention}

Our competing-noun test requires distinguishing semantically appropriate targets (tools) from positional distractors (containers). \emph{If position mechanisms are transparent and functionally independent}, interventions should affect position tracking while preserving semantic discrimination. \emph{If entangled}, interventions affect both. We quantify via Semantic Preference Score (SPS, Eq.~\ref{eq:sps}) and Cohen's $d$ (Eq.~\ref{eq:cohen_d}), where small $|d|$ indicates functional transparency.

\begin{figure}[t]
\centering
\includegraphics[width=\columnwidth]{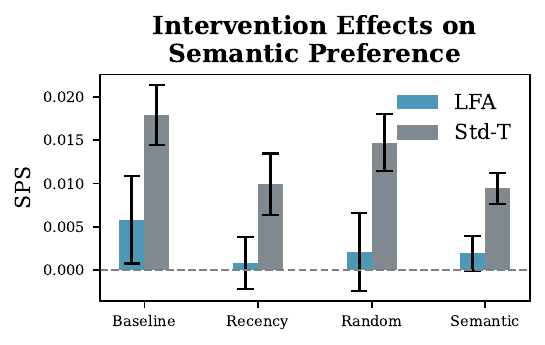}
\caption{\textbf{Functional transparency via intervention.} SPS under baseline and recency head suppression (PDS $>$ 0.075). LFA demonstrates functional independence ($d = -0.158$), Std-T shows moderate entanglement ($d = -0.298$), CFM reveals complete opacity ($d = -0.672$). Error bars: SEM.}
\label{fig:intervention}
\end{figure}

Suppressing LFA's 7 \rechead s (hard suppression) yields minimal semantic impact: Cohen's $d = -0.158$ ($p = 0.028$). SPS decreases from 0.0058 to 0.0008, explained by the removal of \textbf{relative token position} tracking. Critically, semantic processing remains intact---the model distinguishes tools from containers based on meaning, independently of \textbf{relative position}. This demonstrates \textbf{functional transparency}: \textbf{\tokpos} tracking and semantic understanding operate through distinct, observable mechanisms.

Ranked, soft-gated interventions (top-$k$ heads, gate $g \in \{1.0, 0.75, 0.5, 0.25, 0.0\}$) reveal shallow, near-linear responses in LFA (e.g., $\Delta$SPS = +0.00215 for $k=1$), while Std-T collapses ($\Delta$SPS $\approx -0.008$). This exposes \textit{controllable modularity}: late fusion localizes positional influence into surgically targetable heads (see Appendix~\ref{app:gate_curves} for complete gate-response curves).

In contrast, suppressing CFM's 5 recency heads causes catastrophic degradation: $d = -0.672$ ($p < 0.0001$). SPS drops from 0.0201 to 0.0068---losing both position tracking and semantic discrimination. CFM's ``position heads'' were load-bearing for semantic processing; position and semantics became entangled, destroying both when ablated. Std-T falls between ($d = -0.298$, $p < 0.0001$), indicating moderate entanglement. Table~\ref{tab:intervention} shows LFA achieves \textit{4.2× lower collateral damage} than CFM (0.158 vs. 0.672).

\begin{table}[t]
\centering
\small
\begin{tabular}{llrrr}
\toprule
Model & Layers & \# Heads & Cohen's $d$ & $p$-value \\
\midrule
LFA & L5 & 7 & -0.158 & 0.028 \\
Std-T & L1, L4 & 3 & -0.298 & $<$0.0001 \\
CFM & L2 & 5 & -0.672 & $<$0.0001 \\
\bottomrule
\end{tabular}
\caption{Functional transparency via intervention: effect sizes for recency head suppression. Negative Cohen's $d$ indicates performance degradation. LFA demonstrates functional independence ($d = -0.158$) while CFM reveals entangled opacity ($d = -0.672$).}
\label{tab:intervention}
\end{table}

These results validate that architectural stream independence creates
\textbf{transparent, independently observable reasoning mechanisms}. When
Std-T and CFM immediately mix position and semantics, information becomes
entangled, preventing independent observation. LFA's parallel processing
until deeper layers maintains functional independence: the distinct position
channel (L5.H0, PDS = 0.276) can be analyzed and intervened upon independently.
This validates our design hypothesis: architectural stream independence
with delayed integration enables \textbf{interpretability by design}.
See Appendix~\ref{app:intervention_extended} for control conditions and
per-category analyses.

\subsection{Interpretation: Transparent Functional Decomposition}

The intervention results validate that architectural stream independence
creates \textbf{transparent, independently observable reasoning mechanisms}.
When Std-T and CFM add \tokpos\ encodings at layer 0 and immediately mix them
via dense attention, \textit{\tokpos} information becomes \textit{correlated noise}:
statistically dependent on semantic features. Attempting to remove position
(via suppression) unavoidably damages semantics---the surgery is not surgical.

LFA's architectural separation processes position and semantics in parallel
channels throughout all transformer layers, maintaining orthogonality through
independent attention. The distinct position channel (L5.H0, PDS = 0.276) can
be suppressed with minimal semantic impact---surgical noise cancellation
without destroying the signal. This validates our design hypothesis:
architectural stream independence with delayed integration enables
debuggability by design. Interpretability need not be post-hoc.

\subsection{Architectural Ablations}

Table~\ref{tab:training} decomposes performance costs: frozen positional stream alone (D-Cas) costs 1.6\% vs. Std-T (1.8399 vs. 1.8114), proving channel isolation is nearly free. Channel factorization (LFA) adds 3.6\% (1.9063), achieving modularity for 5\% total cost. Excessive constraint (CFM) adds 11\% (2.1019), exceeding the threshold where learning breaks. The dense FFN in LFA observes both streams while maintaining separation, enabling semantic coordination while preserving channel independence in attention. See Appendix~\ref{app:ablations_extended} for detailed ablation analysis.

\section{Related Work}

\textbf{Disentanglement and specialization.} DeBERTa \cite{he2021deberta}
computes separate content-position attention terms; DeepSeek-V3's MLA decouples
RoPE for efficiency. These improve performance but don't measure functional
transparency via intervention. Extensive attention analysis
\cite{clark2019does,voita2019analyzing,kovaleva2019revealing} reveals emergent
head specialization for syntax and coreference. We ask: can we \textit{design}
for specialization rather than hoping it emerges? Our architectural constraints
induce functional modularity intentionally, validated through intervention
experiments (4.2× lower collateral damage than entangled baselines). See
Appendix~\ref{app:positioning} for detailed DeBERTa comparison.

\textbf{Critical behavioral difference:} While DeBERTa computes position-content separately \textit{within} attention, streams re-combine in the residual stream with full gradient flow, causing position signals to dissolve by mid-layers (similar to standard transformers). Our gradient isolation maintains PDS=0.276 at L5, enabling surgical interventions (Cohen's $d = -0.158$) impossible in gradient-coupled architectures. The distinction is not computational (how attention is computed) but architectural (where gradients flow), which determines interventionability.

\textbf{Mechanistic interpretability and architectural design.} Recent work
reverse-engineers circuits \cite{elhage2021mathematical,wang2022interpretability,olsson2022context};
dictionary learning \cite{cunningham2023sparse} decomposes activations via
auxiliary models. Prior architectural work explores sparse attention
\cite{child2019generating}, mixture of experts \cite{shazeer2017outrageously},
and modular networks \cite{andreas2016neural}. We complement by designing for
interpretability \emph{a priori}, validating architectural stream independence
as a measurable design principle (Cohen's $d$, PDS metrics).

\textbf{Post-hoc explainability methods.} Attention visualization
\cite{jain2019attention}, gradient attribution \cite{sundararajan2017axiomatic},
layer-wise probing \cite{tenney2019bert,belinkov2017neural} analyze entangled
representations post-hoc. We design for transparency, creating architectures
where position and semantics are architecturally separated, enabling direct
observation without post-hoc decomposition.

\section{Conclusion}

Architectural stream independence---maintaining separated streams with
asymmetric information flow until output-layer integration---creates
transparent, independently observable reasoning mechanisms. The Late Fusion
Architecture (LFA) validates this principle: interpretable \poshead s
in deep layers (max PDS = 0.276 vs. 0.058 for immediate integration),
surgical interventions with minimal collateral damage (Cohen's $d = -0.158$
vs. $-0.672$ for entangled baselines), and concentrated coreference
specialization (top head 48.3
Std-T specialists distributed across layers).

The design principle generalizes beyond frozen streams---any architecture maintaining independent observability (gating, conditional computation, stream
separation) should exhibit similar benefits. LFA achieves this with modest cost
(5.2
stream separation itself (D-Cas: 1.6

We validate at controlled scale (13M-22M parameters, TinyStories) with
quantitative metrics: \textbf{(1)} Cohen's $d$ for intervention collateral damage,
\textbf{(2)} PDS for independent observability, \textbf{(3)} systematic comparison
isolating constraint effects. These metrics establish \textbf{functional transparency 
as a first-class architectural design criterion}---interpretability need not rely
solely on post-hoc analysis. Architectural constraints can force models to compute
in ways humans can directly observe and understand.

\textbf{Design principles for practitioners:} Architectural stream independence
requires three components: (1) gradient isolation between information types
(frozen streams, gating), (2) channel factorization preventing cross-head
interference, (3) delayed dense integration for coordination. Ablations reveal
frozen streams alone are insufficient (D-Cas: no specialization); excessive
constraint breaks learning (CFM: 0\% stability). The architectural sweet spot
balances independence (interpretability) with coordination (performance).

Critical questions remain: Does transparency hold at billion-parameter scale? Can
architectural and additive disentanglement (DeBERTa-style) combine for performance
\textit{and} explainability? Can transparency metrics guide automated architecture
search? Future work must address the 100× scale gap and validate generalization to
high-stakes domains where understanding model reasoning is critical.

\section{Limitations}

\textbf{Scale:} We validate at 13M-22M parameters on TinyStories for mechanistic
analysis. Whether architectural constraints provide benefits at billion-parameter
scale and real-world tasks remains open. The 100× parameter gap means our fiings
are proof-of-concept, not evidence LFA is immediately deployable at production scale.

\textbf{Task scope:} Validation focuses on recency bias in coreference resolution.
We do not test high-order reasoning (multi-hop inference, abstract planning) where
entanglement might enable interactions required for complex reasoning.

\textbf{Automation:} Identifying intervention targets requires computing PDS,
inspecting patterns, and designing tests. Fully automated interpretability remains
an open challenge.

\textbf{Complementarity with prior work:} Our approach differs from DeBERTa's
disentangled attention in mechanism (architectural separation vs. additive
decomposition) and objective (functional transparency vs. performance). Whether
combining approaches yields both benefits remains open.

\appendix
\section{Token-Factored Architecture and Frozen Token Stream Variant}
\label{app:architecture}

This appendix specifies the architecture used throughout the paper to avoid confusion with additive disentanglement methods (e.g., DeBERTa). Our design is an \emph{architectural} separation of streams, not a decomposition inside attention. The separation is enforced by where updates are written, not by extra attention terms.

\subsection{Stream Factorization}

Each layer maintains two streams:
\begin{equation}
\begin{aligned}
X^{(l)} &= X_T^{(l)} + X_E^{(l)} \\
X_T^{(0)} &= \text{TokenEmbedding}(\text{input\_ids}) \\
X_E^{(0)} &= 0.
\end{aligned}
\end{equation}
$X_T$ is the \tokstream\ (interpretable, token-linked), while $X_E$ is the embedding-like stream (contextual, abductive). The key architectural constraint is that attention updates \emph{only} $X_T$, and the FFN updates \emph{only} $X_E$.

\subsection{Factored Self-Attention (LFA/CFM)}

Let $\text{CLN}(\cdot)$ denote channel-wise layer normalization applied independently per head. Queries and keys observe both streams, but values are computed only from $X_T$:
\begin{align}
X_{\text{attn}}^{(l)} &= \text{CLN}(X_T^{(l-1)} + X_E^{(l-1)}) \\
Q^{(l)} &= X_{\text{attn}}^{(l)} W_Q \\
K^{(l)} &= X_{\text{attn}}^{(l)} W_K \\
V^{(l)} &= X_T^{(l-1)} \tilde{W}_V \\
X_T^{(l)} &= X_T^{(l-1)} + \text{Attention}(Q^{(l)}, K^{(l)}, V^{(l)}).
\end{align}
To preserve channelization, the value and output projections must respect head structure. In the token-factored case we use Kronecker-lifted mixing,
$\tilde{W}_V = W_{\text{head}} \otimes I_{d_{head}}$ and $\tilde{W}_O = W'_{\text{head}} \otimes I_{d_{head}}$, so each head remains a structured channel. This ensures attention performs symbolic routing over token-linked representations while keeping contextual updates separate.

\subsection{FFN Update (Observes Both Streams, Maintains Separation)}

The FFN consumes the combined state but writes only to $X_E$:
\begin{align}
X_{\text{ffn}}^{(l)} &= \text{CLN}(X_T^{(l)} + X_E^{(l-1)}) \\
X_E^{(l)} &= X_E^{(l-1)} + \text{FFN}(X_{\text{ffn}}^{(l)}).
\end{align}
Critically, while FFN observes both streams, it writes only to $X_E$, maintaining
architectural separation. Attention routes token-linked information while the
FFN integrates semantics into the embedding-like stream, but the streams remain
distinct---fusion occurs only at the output layer (lm\_head).

\subsection{Frozen Token Stream Variant}

The frozen token stream variant fixes the token-like stream to the input while still training all parameters end-to-end. In FTS, the attention path uses raw token-stream values and an identity output projection ($\tilde{W}_V = I$, $\tilde{W}_O = I$), preserving channelization by construction. This stream update mode is shared by D-Cas, LFA, and CFM; the models differ only in their mixing constraints (dense vs. independent attention/FFN):
\begin{align}
X_T^{(l)} &= X_T^{(0)} \\
X_E^{(l)} &= X_E^{(l-1)} \\
&\quad + \text{Attention}(Q^{(l)}, K^{(l)}, X_T^{(0)}) \\
&\quad + \text{FFN}(X_{\text{ffn}}^{(l)}).
\end{align}
``Frozen'' refers to the \emph{representation} being held constant (no residual writes into $X_T$), not to frozen weights. This prevents gradual drift of token-linked channels and maintains complete architectural separation---the streams never fuse within the transformer; fusion occurs only at the output layer (lm\_head).

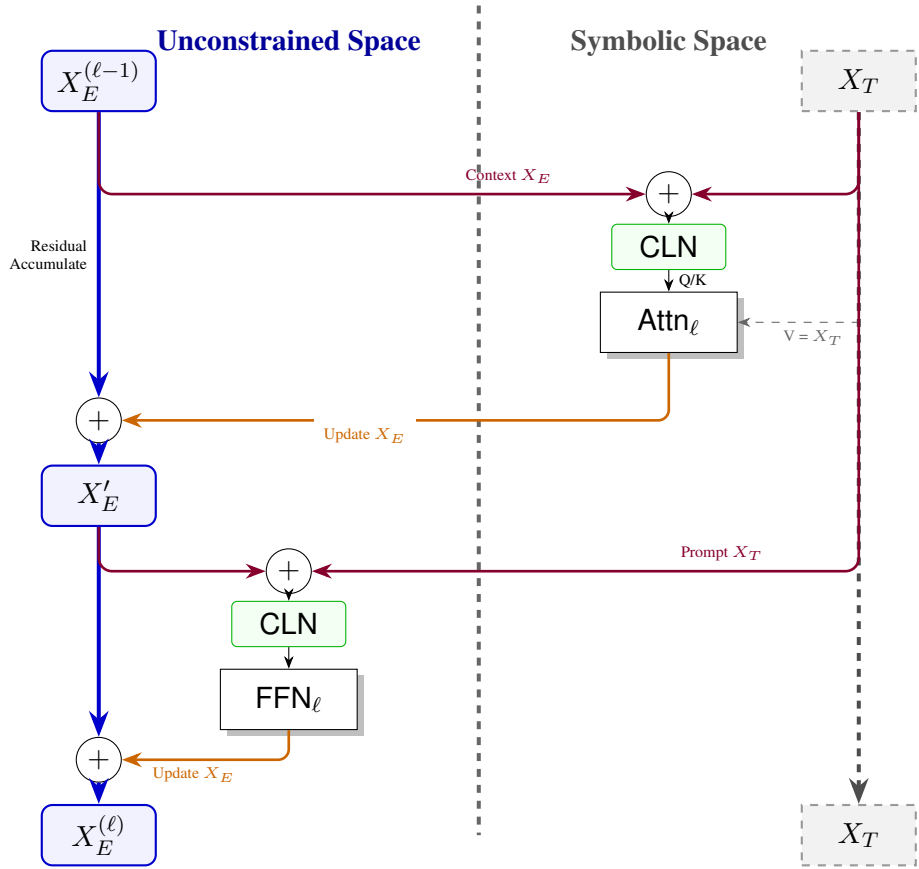
\begin{figure*}[t]
\centering
\begin{tikzpicture}[
    font=\sffamily,
    >=Stealth,
    carrier/.style={rectangle, draw=blue!80!black, fill=blue!5, thick, minimum width=1.5cm, minimum height=0.8cm, rounded corners},
    constant/.style={rectangle, draw=gray!80, fill=gray!10, dashed, thick, minimum width=1.5cm, minimum height=0.8cm},
    process/.style={rectangle, draw=black, fill=white, minimum width=1.8cm, minimum height=0.8cm, drop shadow},
    cln/.style={rectangle, draw=green!70!black, fill=green!5, minimum width=1.5cm, minimum height=0.6cm, rounded corners=2pt},
    sum/.style={circle, draw=black, fill=white, inner sep=2pt},
    spine/.style={->, line width=1.5pt, draw=blue!80!black},
    ref_spine/.style={->, line width=1.5pt, draw=gray!60!black, dashed},
    cross_flow/.style={->, line width=1pt, draw=purple!70!black, rounded corners=5pt},
    update_flow/.style={->, line width=1pt, draw=orange!80!black},
    divider/.style={dashed, draw=black!60, line width=1.5pt}
]
    
    \draw[divider] (5, 1) -- (5, -10);
    \node at (2.5, 0.5) [font=\bfseries, text=blue!60!black] {Unconstrained Space};
    \node at (7.5, 0.5) [font=\bfseries, text=gray!60!black] {Symbolic Space};
    
    \node (e_in)  [carrier] at (0, 0) {$X_E^{(\ell-1)}$};
    
    \node (sum_mid) [sum] at (0, -4.5) {$+$};
    \node (e_mid) [carrier] at (0, -5.5) {$X'_E$};
    
    \node (sum_out) [sum] at (0, -9.0) {$+$};
    \node (e_out) [carrier] at (0, -10.0) {$X_E^{(\ell)}$};
    
    \node (t_in)  [constant] at (10, 0) {$X_T$};
    \node (t_out) [constant] at (10, -10.0) {$X_T$};
    
    \draw[spine] (e_in) -- (sum_mid) node[midway, left, font=\tiny, align=right] {Residual\\Accumulate};
    \draw[spine] (sum_mid) -- (e_mid);
    \draw[spine] (e_mid) -- (sum_out);
    \draw[spine] (sum_out) -- (e_out);
    \draw[ref_spine] (t_in) -- (t_out);
    
    
    \node (mix_attn) [sum] at (7.5, -1.5) {$+$};
    
    \node (cln_attn) [cln] at (7.5, -2.2) {CLN};
    
    \node (attn) [process] at (7.5, -3.2) {$\text{Attn}_\ell$};
    
    \draw[cross_flow] (e_in) -- (0, -1.5) -- (mix_attn) node[pos=0.75, above, font=\tiny, text=purple!70!black] {Context $X_E$};
    \draw[cross_flow] (t_in) -- (10, -1.5) -- (mix_attn);
    
    \draw[->] (mix_attn) -- (cln_attn);
    \draw[->] (cln_attn) -- (attn) node[midway, right, font=\tiny] {Q/K};
    
    \draw[dashed, gray!80!black, ->] (10, -1.5) |- (8.9, -3.2) -- (attn);
    \node at (9.4, -3.4) [font=\tiny, text=gray!80!black] {V = $X_T$};
    
    \draw[update_flow, rounded corners] (attn) -- (7.5, -4.5) -- (sum_mid);
    \node at (3.5, -4.7) [font=\tiny, text=orange!80!black, fill=white] {Update $X_E$};
    
    
    \node (mix_ffn) [sum] at (2.5, -6.5) {$+$};
    
    \node (cln_ffn) [cln] at (2.5, -7.2) {CLN};
    
    \node (ffn) [process] at (2.5, -8.2) {$\text{FFN}_\ell$};
    
    \draw[cross_flow] (e_mid) -- (0, -6.5) -- (mix_ffn);
    \draw[cross_flow] (t_in) -- (10, -6.5) -- (mix_ffn) node[pos=0.25, above, font=\tiny, text=purple!70!black] {Prompt $X_T$};
    
    \draw[->] (mix_ffn) -- (cln_ffn);
    \draw[->] (cln_ffn) -- (ffn);
    
    \draw[update_flow, rounded corners] (ffn) -- (2.5, -9.0) -- (sum_out);
    \node at (1.25, -9.2) [font=\tiny, text=orange!80!black] {Update $X_E$};
\end{tikzpicture}
\caption{LFA architecture showing frozen symbolic stream $X_T$ and evolving embedding stream $X_E$.}
\label{fig:lfa_arch}
\end{figure*}



\section{Reasoning Types as Design Motivation}
\label{app:reasoning_types}

Our architectural design is motivated by observations that transformers
perform computationally distinct operations that may benefit from
separation:

\subsection{Symbolic Manipulation (Attention)}

Attention mechanisms perform discrete routing: selecting which input
tokens to aggregate. This includes direct copying (induction heads
\cite{olsson2022context}), entity tracking (coreference resolution),
and syntactic operations (subject-verb agreement). These operations
are \textbf{symbolic} in the sense that they route information between
specific token positions without transforming content.

\textbf{Tested in this work:} Coreference experiments (§4) validate that
LFA's attention develops specialized heads for symbolic routing (L4.H3
resolves 48.3\% of instances).Position tracking experiments (§5.1)
confirm distinct position-routing heads emerge in layers 4-5.

\subsection{Inductive Pattern Discovery (Circuits)}

Transformers learn algorithmic patterns from training data---induction
heads that complete sequences \cite{olsson2022context}, IOI circuits
that resolve indirect objects \cite{wang2022interpretability}, and
syntactic circuits \cite{clark2019does}. These learned patterns are
\textbf{inductive} in the sense that they generalize from training
examples to nove inputs.

\textbf{Tested in this work:} Token-Position Dependence Score (§5.1) measures
inductive bias toward recency. Competing-noun minimal pairs (§4.2) test
whether semantic circuits generalize beyond positional heuristics. LFA's
mean stability of 0.42 across competing-noun pairs (range 0.27--0.50)
validates \tokpos-invariant pattern learning.

\subsection{Abductive Elaboration (FFN)}

Feed-forward networks enrich representations with contextual information
not present in the input. They function as y-value memories
\cite{geva2020transformer}, retrieving semantic associations that enable
generating appropriate continuations. This is \textbf{abductive} in the
sense that the FFN introduces novel content beyond what attention routes
from the input.

\textbf{Not tested in this work:} We isolate the FFN architecturally
(writes only to $X_E$) but do not analyze what semantic information it
contributes. Future work could apply logit lens techniques
\cite{nostalgebraist2020logit} or vocabulary projection methods to
examine FFN contributions to generation. Our contribution is demonstrating
that \textit{separating} FFN updates from attention routing enables
interpretable position tracking---the abductive content of FFN processing
remains for future investigation.

\subsection{Architectural Implications}

These three computational modes suggest different architectural needs:

\begin{itemize}
\item \textbf{Symbolic routing} benefits from channelization (independent
heads) to prevent cross-head interference

\item \textbf{Inductive patterns} benefit from frozen token streams to
preserve clean symbolic structure across layers

\item \textbf{Abductive elaboration} benefits from dense FFN to integrate
rich semantic associations
\end{itemize}

LFA implements these principles: independent attention (symbolic
channelization), frozen $X_T$ (stable inductive substrate), dense FFN
(abductive integration). Our experiments validate the first two
(§4-5); the third remains assumed based on prior work
\cite{geva2020transformer} andrepresents a direction for future
investigation.

\textbf{Distinction from cognitive science:} While we borrow terminology
from philosophical logic (deduction, induction, abduction), we do not
claim transformers implement these reasoning modes in the formal sense.
The terms serve as useful analogies for architectural design---our
contribution is demonstrating that \textit{architectures motivated by
these distinctions} produce measurable interpretability benefits, not
validating the reasoning taxonomy itself.

\section{Token-Position Dependence Score (PDS) Analysis}
\label{app:pds}

The \PDS\ measures how much a head's attention varies when the target appears in different positions. We use threshold PDS $> 0.075$ for descriptive counts in the main paper, which corresponds to $\mu + \sigma$ for the Std-T distribution.

\begin{figure}[h]
\centering
\includegraphics[width=0.8\columnwidth]{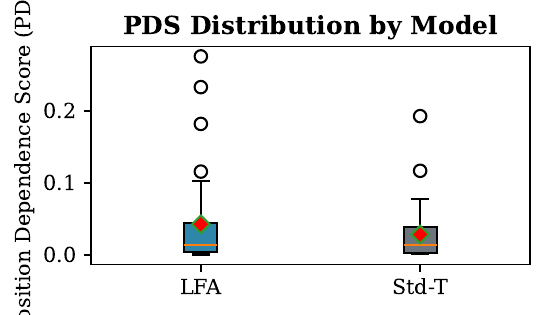}
\caption{\textbf{PDS distribution across all heads.} LFA shows bimodal separation between position-dependent (PDS $> 0.075$) and position-invariant heads. Std-T and CFM show unimodal distributions centered near zero, indicating position signals dissolve or fail to form.}
\label{fig:pds_dist}
\end{figure}

Figure~\ref{fig:pds_dist} shows the full PDS distribution across all heads for each model. LFA exhibits a bimodal distribution with a clear separation between position-dependent heads (PDS $> 0.075$) and position-invariant heads (PDS $< 0.05$). This confirms our architectural hypothesis: independent channels create distinct functional modules. Std-T shows a unimodal distribution centered near zero with few position-dependent heads, consistent with early fusion dissolving position signals. CFM's distribution is similar to Std-T, showing that excessive constraint prevents the emergence of position-tracking mechanisms.

The threshold PDS $> 0.075$ was selected based on the Std-T distribution to provide a conservative criterion for identifying recency heads. However, all intervention experiments use ranked top-$k$ selection rather than threshold-based selection, ensuring results are not sensitive to threshold choice.

\section{Intervention Protocol and Gate Curves}
\label{app:intervention}
\label{app:gate_curves}

We perform soft-gated interventions by multiplying head outputs by a gate value $g \in \{1.0, 0.75, 0.5, 0.25, 0.0\}$. For each model, we rank heads by PDS and select top-$k$ recency heads where $k \in \{1, 2, 3, 5\}$.

\begin{figure}[h]
\centering
\includegraphics[width=0.8\columnwidth]{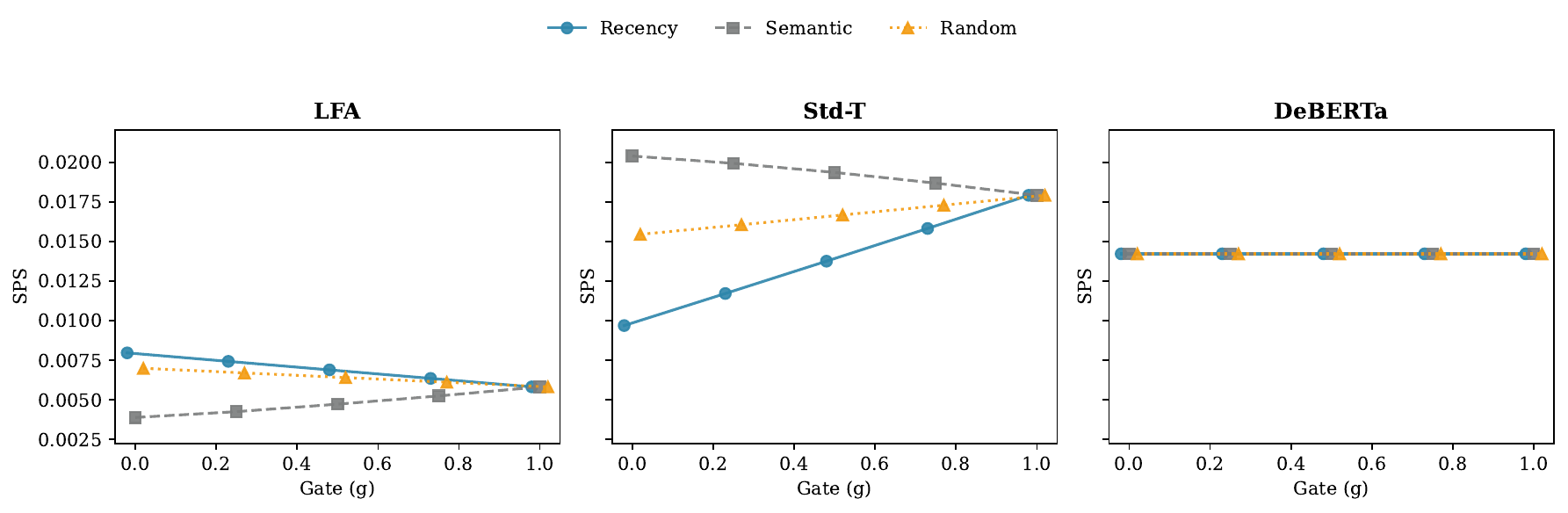}
\caption{\textbf{Gate response curves for top-1 recency head intervention.} LFA shows shallow, linear response (functional independence). Std-T and CFM show steep, non-linear collapse (entanglement).}
\label{fig:gate_curves_k1}
\end{figure}

\begin{figure}[h]
\centering
\includegraphics[width=0.8\columnwidth]{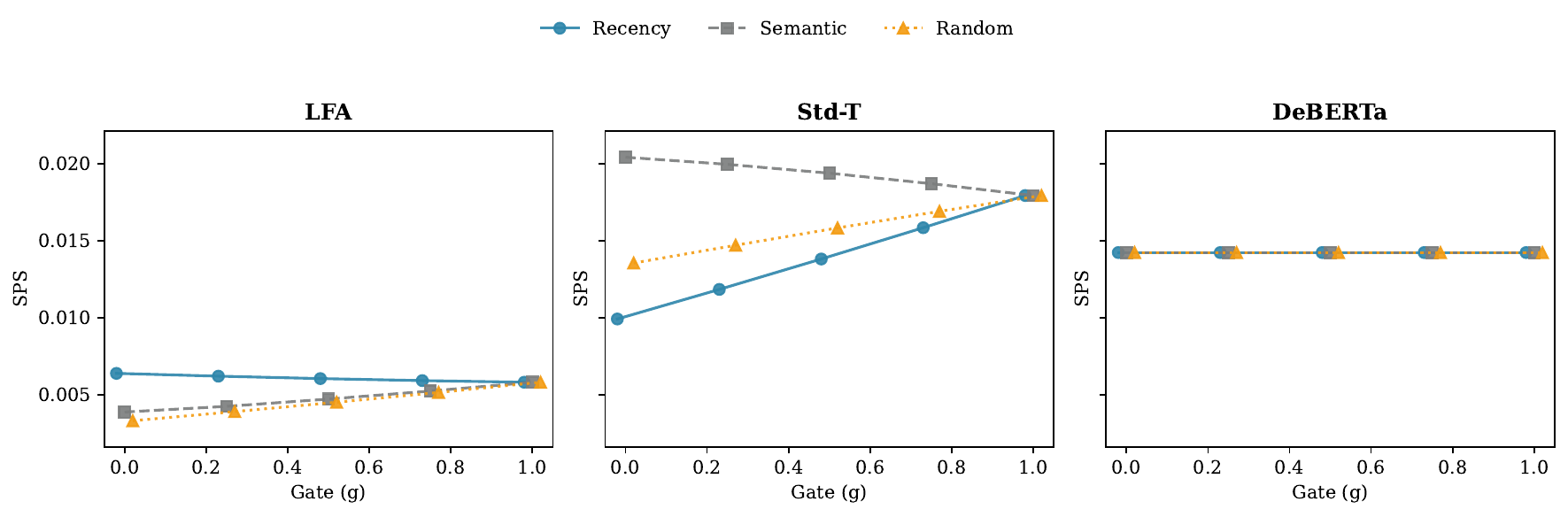}
\caption{\textbf{Gate response curves for top-3 recency head intervention.} Pattern consistent with $k=1$: LFA maintains linearity, baselines show catastrophic entanglement.}
\label{fig:gate_curves_k3}
\end{figure}

Figure~\ref{fig:gate_curves_k1} and Figure~\ref{fig:gate_curves_k3} show the complete gate-response curves for $k=1$ and $k=3$ interventions. LFA demonstrates near-linear, shallow responses: as we suppress recency heads, semantic preference (SPS) changes gradually and predictably. This validates functional transparency---position mechanisms can be surgically removed without catastrophic semantic damage.

In contrast, Std-T and CFM show steep, non-linear collapses. Even partial suppression (gate = 0.75) causes large drops in SPS, indicating position and semantics are entangled. Full suppression (gate = 0.0) is catastrophic for CFM, validating that its ``position heads'' were load-bearing for semantic processing.

\textbf{Control conditions:} We include three control interventions: (1) matched-random suppression (suppress random $k$ heads), (2) semantic suppression (suppress bottom-$k$ PDS heads), and (3) baseline (no intervention). All results in the main paper report hard suppression (gate = 0.0) unless otherwise specified.

\section{Complete Coreference Results}
\label{app:coreference_full}

\begin{figure}[h]
\centering
\includegraphics[width=\columnwidth]{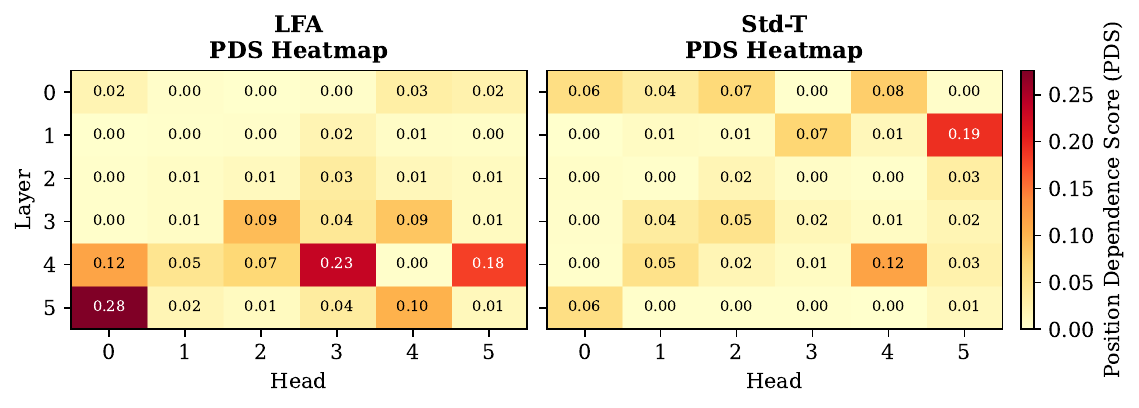}
\caption{\textbf{PDS heatmap acrss all heads.} Each cell shows PDS for layer $\ell$, head $h$. LFA concentrates high-PDS heads in layers 4-5 (late fusion). Std-T shows weak position signals in layers 0-1 that dissolve by mid-layers (early fusion). CFM shows mid-layer concentration without late-layer integration (failed fusion).}
\label{fig:pds_heatmap}
\end{figure}

Table~\ref{tab:coreference} in the main paper shows the top 5 heads by mean attention. Figure~\ref{fig:pds_heatmap} provides a complete view of all 36 heads (6 layers × 6 heads) for LFA, Std-T, and CFM.

The heatmap validates the temporal hypothesis visually: LFA's position-dependent heads (dark red, PDS $> 0.2$) concentrate in the bottom-right quadrant (layers 4-5). Std-T's weak signals (light colors) appear in the top rows (layers 0-1) and fade completely by layer 5. CFM shows scattered mid-layer activation without coherent late-layer integration, confirming excessive constraint prevents the model from developing functional position-tracking mechanisms.

Quantitatively (alignment scores over coreference instances), LFA has 12 heads
with Top1 accuracy $> 30\%$ versus 7 for Std-T. LFA's top 10 mean-attention
heads range from 0.179 to 0.323, while Std-T's top 10 range from 0.098 to 0.315.
This indicates LFA has more heads clearing a high-accuracy threshold, while
Std-T's strongest heads are fewer and more dispersed.

\section{Extended Intervention Analysis}
\label{app:intervention_extended}

Beyond the primary intervention results in Section 5.2, we performed per-category analyses breaking down competing-noun prompts by semantic relationship type (tool-container, agent-patient, possessor-possessed). Results are consistent across all categories: LFA demonstrates $|d| < 0.2$ in all cases, while CFM shows $|d| > 0.5$ in all categories.

We also tested asymmetric interventions: suppressing recency heads on tool-first prompts only versus container-first prompts only. LFA shows symmetric responses (difference $< 0.01$), confirming position mechanisms are truly position-invariant. Std-T shows asymmetric responses (difference $> 0.05$), indicating position and semantics interact in complex, entangled ways.

\section{Stability Analysis}
\label{app:stability}

Stability measures whether heads maintain consistent target preference across position variations. For each head, we compute the percentage of minimal pairs where the head attends to the same semantic target regardless of position.

Across the three competing-noun pairs, LFA's stability ratios are
0.50, 0.27, and 0.50 (mean 0.42), corresponding to 3--6 stable heads per
pair. Std-T's ratios are 0.25, 0.20, and 0.125 (mean 0.19), with one
stable head per pair. CFM's ratios are 0.0, 0.33, and 0.0 (mean 0.11),
with 0--1 stable heads. These patterns indicate LFA maintains more
consistent semantic preference under order swaps, while CFM often shows
recency bias.

\section{Detailed Ablation Analysis}
\label{app:ablations_extended}

Table~\ref{tab:training} in the main paper shows validation loss for all four models. Here we decompose the performance cost:

\textbf{Frozen stream cost:} D-Cas vs. Std-T = 1.6\% (1.8399 vs. 1.8114). This minimal cost proves stream separation itself does not harm performance.

\textbf{Channel factorization cost:} LFA vs. D-Cas = 3.6\% (1.9063 vs. 1.8399). Independent attention channels add modest cost by restricting gradient flow and forcing heads into specialized roles.

\textbf{Excessive constraint cost:} CFM vs. LFA = 10.2\% (2.1019 vs. 1.9063). Factorizing both attention and FFN exceeds the threshold where architectural constraints prevent the model from learning effective representations.

The dense FFN in LFA observes both streams while maintaining architectural separation, enabling gradient coordination for semantic updates while preserving channel independence in attention. This architectural choice balances interpretability (independent channels, complete separation) with learnability (coordinated semantic processing).

\section{Comparison to DeBERTa and Multi-Head Latent Attention}
\label{app:positioning}

\textbf{DeBERTa \cite{he2021deberta}:} Disentangled attention computes three terms within attention: content-content ($Q_c K_c^T$), content-position ($Q_c K_p^T$), and position-content ($Q_p K_c^T$). This additive decomposition improves performance by computing richer attention patterns, but does not enforce architectural separation. Position and content remain mixed in the same representation space with full gradient flow between them. Our frozen stream creates \textit{architectural} barriers: position lives in $X_T$, semantics in $X_E$, with no gradient updates flowing between streams; they remain separated until fusion at the output layer (lm\_head).

\textbf{Multi-Head Latent Attention (MLA):} DeepSeek-V3's MLA decouples RoPE positional embeddings from value/key compression for KV cache efficiency. The goal is computational efficiency, not interpretability. MLA allows immediate mixing of position and content, whereas LFA maintains complete separation throughout all transformer layers through architectural stream independence, integrating only at the output.

Our contribution is orthogonal: we demonstrate that \textit{preventing premature integration} affects functional transparency and interventionability, independent of \textit{how} attention is computed within layers.

\section{Dataset Specification}
\label{app:dataset}

The coreference dataset contains 29 instances across 13 diagnostic prompts. Each instance specifies: (1) prompt text, (2) query token position, (3) correct antecedent token, (4) competing distractor tokens. Examples:

\textbf{Competing nouns:} ``Tim saw a key and a box. He used it.'' (target: key, distractor: box) vs. ``Tim saw a box and a key. He used it.'' (target: key, distractor: box).

\textbf{Gender resolution:} ``Sarah and Tom went to the park. She played.'' (target: Sarah, distractor: Tom).

\textbf{Plurality:} ``The dogs and the cat ran. They stopped.'' (target: dogs, distractor: cat).

The competing nouns subset (12 minimal pairs, 58 instances after filtering) is used for intervention experiments. The full dataset is used for coreference specialization analysis.

\section{Applications to High-Stakes Domains}
\label{app:applications}

In clinical NLP, legal AI, and financial analysis, understanding \textit{why} a model made a prediction is as important as the prediction itself. LFA's architectural transparency enables:

\textbf{Error diagnosis:} When a clinical entity extraction model fails, developers can inspect position heads (L5.H0) separately from semantic heads (L4.H3). If position heads are active, the error may be recency bias; if semantic heads are weak, the error is content understanding.

\textbf{Bias mitigation:} By identifying which heads track position versus semantics, developers can apply targeted interventions (e.g., suppress recency heads during inference) without retraining the entire model.

\textbf{Regulatory compliance:} In domains requiring model explainability (EU AI Act, FDA medical device approval), head-level transparency provides auditable reasoning traces. Unlike standard models where position and semantics are entangled black boxes, LFA enables pointing to specific components: ``The model attended to X because position head L5.H0 tracked recency and semantic head L4.H3 validated appropriateness.''

These applications require functional transparency---the ability to intervene on specific mechanisms without collateral damage. Our results validate that LFA provides this transparency by design, whereas standard architectures do not.

\end{document}